\documentclass[10pt,twocolumn,letterpaper]{article}

\usepackage{cvpr}
\usepackage{times}
\usepackage{epsfig}
\usepackage{graphicx}
\usepackage{amsmath}
\usepackage{amssymb}

\usepackage{caption}
\usepackage{subfig}
\usepackage{array}
\usepackage{booktabs}
\usepackage{multirow}

\usepackage[breaklinks=true,bookmarks=false,bookmarksnumbered=true]{hyperref}
\hypersetup{
     colorlinks = true,
     urlcolor = magenta
     }

\cvprfinalcopy 

\def\cvprPaperID{1494} 

\ifcvprfinal\fi
\begin{document}


\title{Detecting Attended Visual Targets in Video}

\author{Eunji Chong$^{1}$\hspace{10mm} Yongxin Wang$^{2}$\hspace{10mm} Nataniel Ruiz$^{3}$\hspace{10mm} James M. Rehg$^{1}$\\
\vspace{-4mm}
\\
{\small $^{1}$Georgia Institute of Technology\hspace{10mm} $^{2}$Carnegie Mellon University\hspace{10mm} $^{3}$Boston University}\\
{\tt\small \{eunjichong,rehg\}@gatech.edu, 
\tt\small yongxinw@andrew.cmu.edu, 
\tt\small nruiz9@bu.edu}
\\
\url{https://github.com/ejcgt/attention-target-detection}
\vspace{-7mm}
}

\twocolumn[{%
\renewcommand\twocolumn[1][]{#1}%
\maketitle
\begin{center}
    \centering
    \includegraphics[width=1\textwidth]{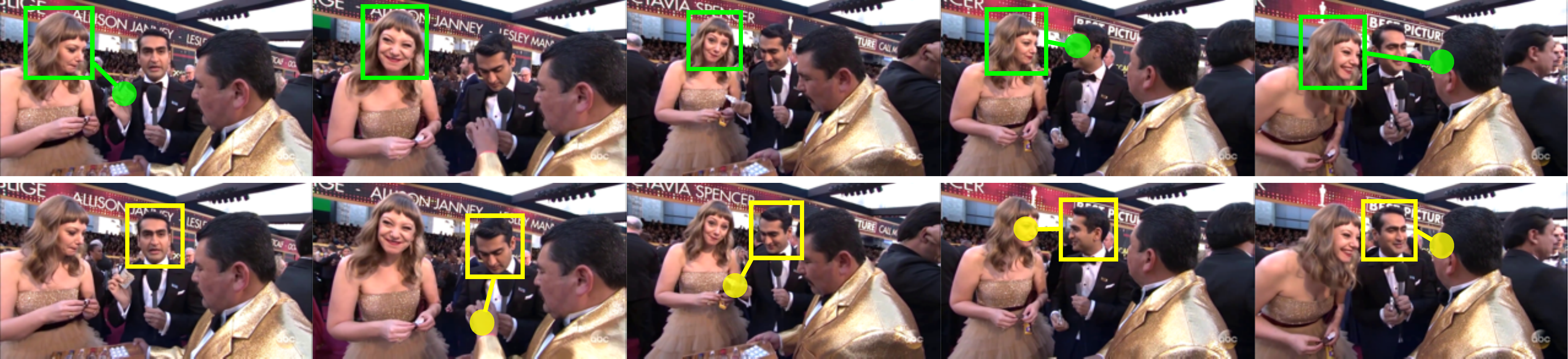}
    \captionof{figure}{\textbf{Visual attention target detection over time.} We propose to solve the problem of identifying gaze targets in video. The goal of this problem is to predict the location of visually attended region (circle) in every frame, given a track of an individual's head (bounding box). It includes the cases where such target is out of frame (row-col: 1-2, 1-3, 2-1), in which case the model should correctly infer its absence.}
    \label{fig:overview}
\end{center}%
}]

\maketitle
\thispagestyle{empty}

\begin{abstract}
   We address the problem of detecting attention targets in video. Our goal is to identify where each person in each frame of a video is looking, and correctly handle the case where the gaze target is out-of-frame. Our novel architecture models the dynamic interaction between the scene and head features and infers time-varying attention targets. We introduce a new annotated dataset, VideoAttentionTarget, containing complex and dynamic patterns of real-world gaze behavior. Our experiments show that our model can effectively infer dynamic attention in videos. In addition, we apply our predicted attention maps to two social gaze behavior recognition tasks, and show that the resulting classifiers significantly outperform existing methods. We achieve state-of-the-art performance on three datasets: GazeFollow (static images), VideoAttentionTarget (videos), and VideoCoAtt (videos), and obtain the first results for automatically classifying clinically-relevant gaze behavior without wearable cameras or eye trackers.
\end{abstract}

\section{Introduction}
Gaze behavior is a critically-important aspect of human social behavior, visual navigation, and interaction with the 3D environment~\cite{kleinke1986gaze,land2009looking}. While monitor-based and wearable eye trackers are widely-available, they are not sufficient to support the large-scale collection of naturalistic gaze data in contexts such as face-to-face social interactions or object manipulation in 3D environments. Wearable eye trackers are burdensome to participants and bring issues of calibration, compliance, cost, and battery life. 

Recent works have demonstrated the ability to estimate the gaze target directly from images, with the potential to greatly increase the scalability of naturalistic gaze measurement. A key step in this direction was the work by Recasens \etal~\cite{recasens2015they}, which demonstrated the ability to detect the attention target of each person within a single image. This approach was extended in~\cite{chong2018connecting} to handle the case of out-of-frame gaze targets. Other related works include~\cite{recasens2017following,saran2018human,lian2018believe,zhao2019learning,guan2019enhance}. These approaches are attractive because they can leverage head pose features, as well as the saliency of potential gaze targets, in order to resolve ambiguities in gaze estimation. 

This paper develops a spatiotemporal approach to gaze target prediction which models the dynamics of gaze from video data. Fig~\ref{fig:overview} illustrates our goal: For each person in each video frame we estimate where they are looking, including the correct treatment of out-of-frame gaze targets. By identifying the visually-attended region in every frame, our method produces a dense measurement of a person's natural gaze behavior. Furthermore, this approach it has the benefit of linking gaze estimation to the broader tasks of action recognition and dynamic visual scene understanding. 

An alternative to the dynamic prediction of gaze targets is to directly classify specific categories or patterns of gaze behaviors from video~\cite{marin2014detecting,palmero2018automatic,marin2019laeo,fan2018inferring,Sumer_2020_WACV,Fan_2019_ICCV}. This approach treats gaze analysis as an action detection problem, for actions such as mutual gaze~\cite{marin2014detecting,palmero2018automatic,marin2019laeo} or shared attention to an object~\cite{fan2018inferring,Sumer_2020_WACV}. While these methods have the advantage of leveraging holistic visual cues,
they are limited by the need to pre-specify and label the target behaviors. In contrast, our approach of predicting dense gaze targets provides a flexible substrate for modeling domain-specific gaze behaviors, such as the assessments of social gaze used in autism research~\cite{mundy2003early,bryson2008autism}.

A key challenge in tackling the dynamic estimation of gaze targets in video is the lack of suitable datasets containing ground truth gaze annotations in the context of rich, real-world examples of complex time-varying gaze behaviors. We address this challenge by introducing the \emph{VideoAttentionTarget} dataset, which contains 1,331 video sequences of annotated dynamic gaze tracks of people in diverse situations.

Our approach to spatiotemporal gaze target prediction has two parts. First, we develop a novel spatial reasoning architecture to improve the accuracy of target localization. The architecture is composed of a scene convolutional layer that is regulated by the head convolutional layer via an attention mechanism~\cite{BahdanauCB14}, such that the model focuses on the scene region that the head is oriented to. The spatial module improves the state-of-the-art result on the GazeFollow benchmark by a considerable margin. Second, we extend the model in the temporal dimension through the addition of ConvLSTM networks. This model outperforms multiple baselines on our novel VideoAttentionTarget dataset. The software, models and dataset are made freely-available for research purposes.

We further demonstrate the value of our approach by using the predicted heatmap from our model for social gaze recognition tasks. Specifically, we experimented on two tasks: 1) Automated behavioral coding of the social gaze of young children in an assessment task, and 2) Detecting shared attention in social scenes. In the first experiment, our heatmap features were found to be the most effective among multiple baselines for attention shift detection. In the second experiment, our approach achieved state-of-the-art performance on the VideoCoAtt dataset~\cite{fan2018inferring}. Both results validate the feasibility and effectiveness of leveraging our gaze target prediction model for gaze behavior recognition tasks. 
This paper makes the following contributions:
\begin{itemize}
    \item A novel spatio-temporal deep learning architecture that learns to predict dynamic gaze targets in video
    \item A new \emph{VideoAttentionTarget} dataset, containing dense annotations of attention targets with complex patterns of gaze behavior
    \item Demonstration that our model's predicted attention map can achieve state-of-the art results on two social gaze behavior recognition tasks
\end{itemize}

\begin{figure*}[ht]
    \centering
    \subfloat[Example frames and annotations]{\includegraphics[width=0.84\textwidth,height=7.5cm]{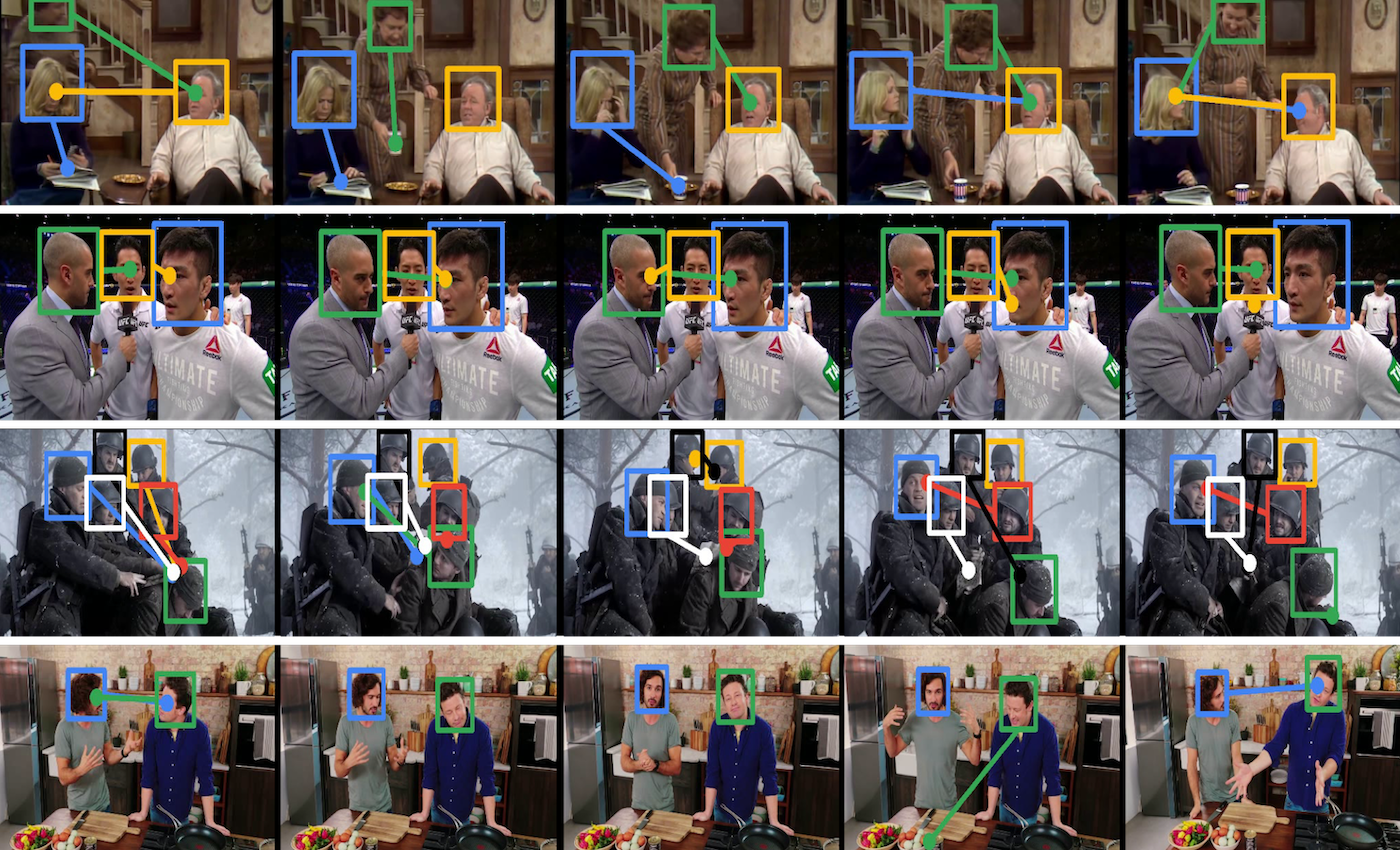}\label{fig:dataset_img}}
    \hfill
    \subfloat[Dataset statistics]{\includegraphics[width=0.15\textwidth,height=7.5cm]{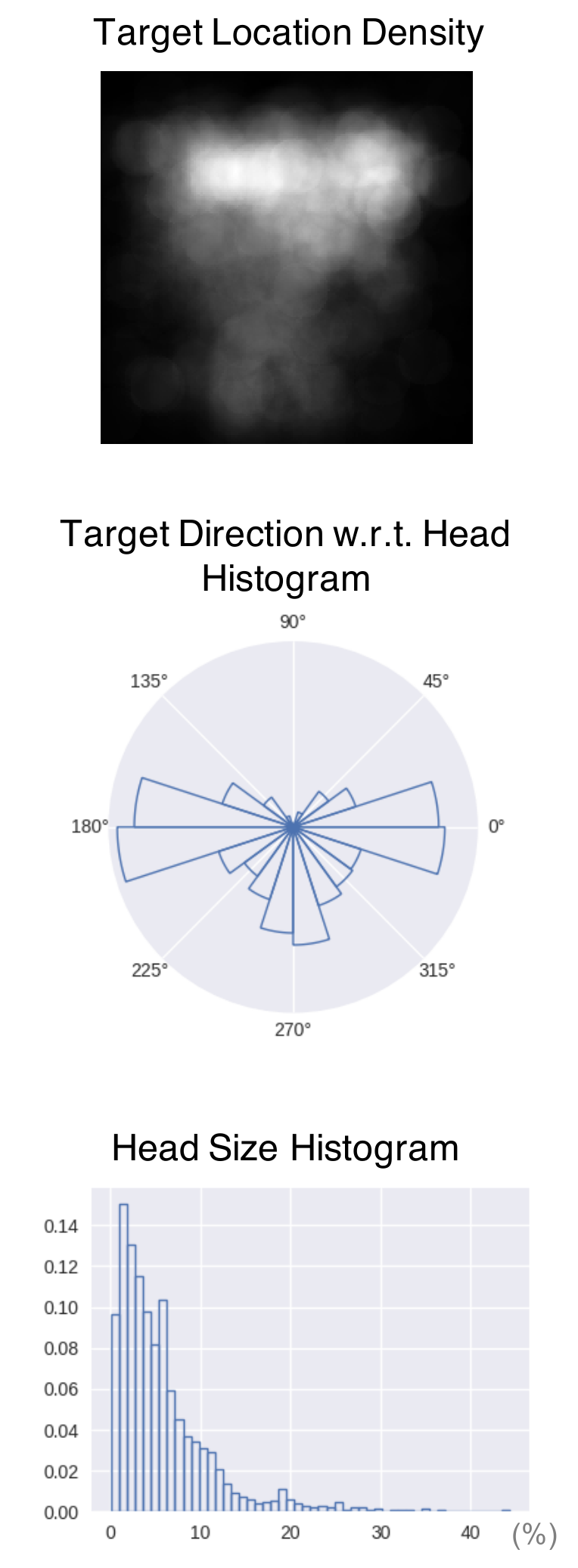}\label{fig:dataset_stat}}
    \caption{\textbf{Overview of novel \emph{VideoAttentionTarget} dataset} (a) Example sequences illustrating the per-frame annotations of each person (bounding box) and their corresponding gaze target (solid dot). (b) Annotation statistics: top - annotated gaze target location distribution in image coordinates, middle - histogram of directions of gaze targets relative to the head center, bottom - histogram of head sizes measured as the ratio of the bounding box area to the frame size.}
    \label{fig:dataset_showcase}
    \vspace{-3mm}
\end{figure*}

\section{Related Work}
We organize the related work into three areas: gaze target prediction, gaze behavior recognition, and applications to social gaze analysis. Our focus is gaze target prediction, but we also provide results for behavior recognition in a social gaze setting (see Secs.~\ref{sec:app1} and~\ref{sec:app2}).

\paragraph{Gaze Target Prediction }
One key distinction among previous works on gaze target prediction is whether the attention target is located in a 2D image~\cite{recasens2015they,recasens2017following,saran2018human,chong2018connecting,lian2018believe,zhao2019learning,guan2019enhance} or 3D space~\cite{ba2008recognizing, masse2017tracking, wei2018and, Brau_2018_ECCV, MasseLMH19}. Our work addresses the 2D case, which we review in more detail; Authors of~\cite{recasens2015they} were among the first to demonstrate how a deep model can learn to find the gaze target in the image. Saran \etal~\cite{saran2018human} adapt the method of~\cite{recasens2015they} to a human-robot interaction task. Chong \etal~\cite{chong2018connecting} extends the approach of~\cite{recasens2015they} to address out-of-frame gaze targets by simultaneously learning gaze angle and saliency. Within-frame gaze target estimation can be further enhanced by considering different scales~\cite{lian2018believe}, body pose~\cite{guan2019enhance} and sight lines~\cite{zhao2019learning}. A key difference between these works and our approach is that we explicitly model the gaze behavior over time and report results for gaze target prediction in video while considering out-of-frame targets.

Our problem formulation and network architecture are most closely-related to~\cite{chong2018connecting}. In addition to the temporal modeling, three other key differences from~\cite{chong2018connecting} are 1) that we do not supervise with gaze angles and not require auxiliary datasets; 2) therefore we greatly simplify the training process; and 3) we present an improved spatial architecture. In terms of architecture, 1) we use head features to regulate the spatial pooling of the scene image via an attention mechanism; 2) we use a head location map instead of one-hot position vector; and 3) we use deconvolutions instead of a grid output to produce a fine-grained heatmap. Our experiments show that these innovations result in improved performance on GazeFollow (\ie for static images, see Table~\ref{tab:eval_gazefollow}) and on our novel video attention dataset (see Table~\ref{tab:main_quantitative}).


The work of~\cite{recasens2017following} shares our goal of inferring gaze targets from video. In contrast to our work, they address the case where the gaze target is primarily visible at a later point in time, after the camera pans or there is a shot change. While movies commonly include such indirect gaze targets, they are rare in the social behavior analysis tasks that motivate this work (see Fig.~\ref{fig:shift_maps}). Our work is complementary to~\cite{recasens2017following}, in that our model infers per-frame gaze targets.

Several works address the inference of 3D gaze targets~\cite{ba2008recognizing, masse2017tracking, MasseLMH19}. In this setting, the identification of an out-of-frame gaze target can be made by relying on certain assumptions about the scene, such as the target object's location or its motion, or by using a joint learning framework informed by the task~\cite{wei2018and} or target location~\cite{Brau_2018_ECCV}.



\paragraph{Gaze Behavior Recognition }
An alternative to inferring the target gaze location is to directly infer a gaze-related behavior of interest. For example, several approaches have been developed to detect if two people are looking at each other~\cite{marin2014detecting,palmero2018automatic,marin2019laeo}, or to detect if more than two people are looking at a common target~\cite{fan2018inferring, Sumer_2020_WACV}. In addition, the 3D detection of socially-salient regions has been investigated using an egocentric approach~\cite{Park2012}. Recently, Fan \etal~\cite{Fan_2019_ICCV} addressed the problem of recognizing atomic-level gaze behavior when human gaze interactions are categorized into six classes such as avert, refer, and follow.

In contrast to approaches that directly infer gaze behavior, our method provides a dense mid-level representation of attention for each person in a video. Thus our approach is complementary to these works, and we demonstrate in Secs.~\ref{sec:app1} and~\ref{sec:app2} that our gaze representation has utility for gaze behavior classification.

\paragraph{Social Gaze Detection in Clinical Settings }
One motivation for our work is the opportunity for automated measurements of gaze behavior to inform research and clinical practice in understanding and treating developmental conditions such as autism~\cite{Rehg2014,hashemi2014computer}. In this setting, automated analysis can remove the burden of laborious gaze coding that is commonplace in autism research, and enable a more fine-grained analysis of gaze behavior in clinical populations. Prior work in this area has leveraged the ability to analyze head orientation~\cite{ruiz2018fine,gu2017dynamic,chong2017visual} to infer children's attention and have developed solutions for specific settings~\cite{pusiol2014discovering,heath2018detecting,Campbell2018}. Prior works have also addressed the detection of eye contact and mutual gaze in the context of dementia care~\cite{Mitsuzumi2017DEEPEC, nakazawa2019first} and autism~\cite{ye2015detecting, chong2017detecting,chong2020detection}. Other work has analyzed mutual gaze in group interactions for inferring rapport~\cite{muller2018robust}. In contrast to these works, our focus is to first develop a general approach to gaze target identification in video, and then explore its utility in estimating clinically-important social behaviors during face-to-face interactions between an adult examiner and a child. We believe are the first to present results (in Sec.~\ref{sec:app1}) for automatically detecting clinically-meaningful social gaze shifts without a wearable camera or an eye tracker.

\begin{figure*}[ht]
    \centering
    \includegraphics[width=1\textwidth]{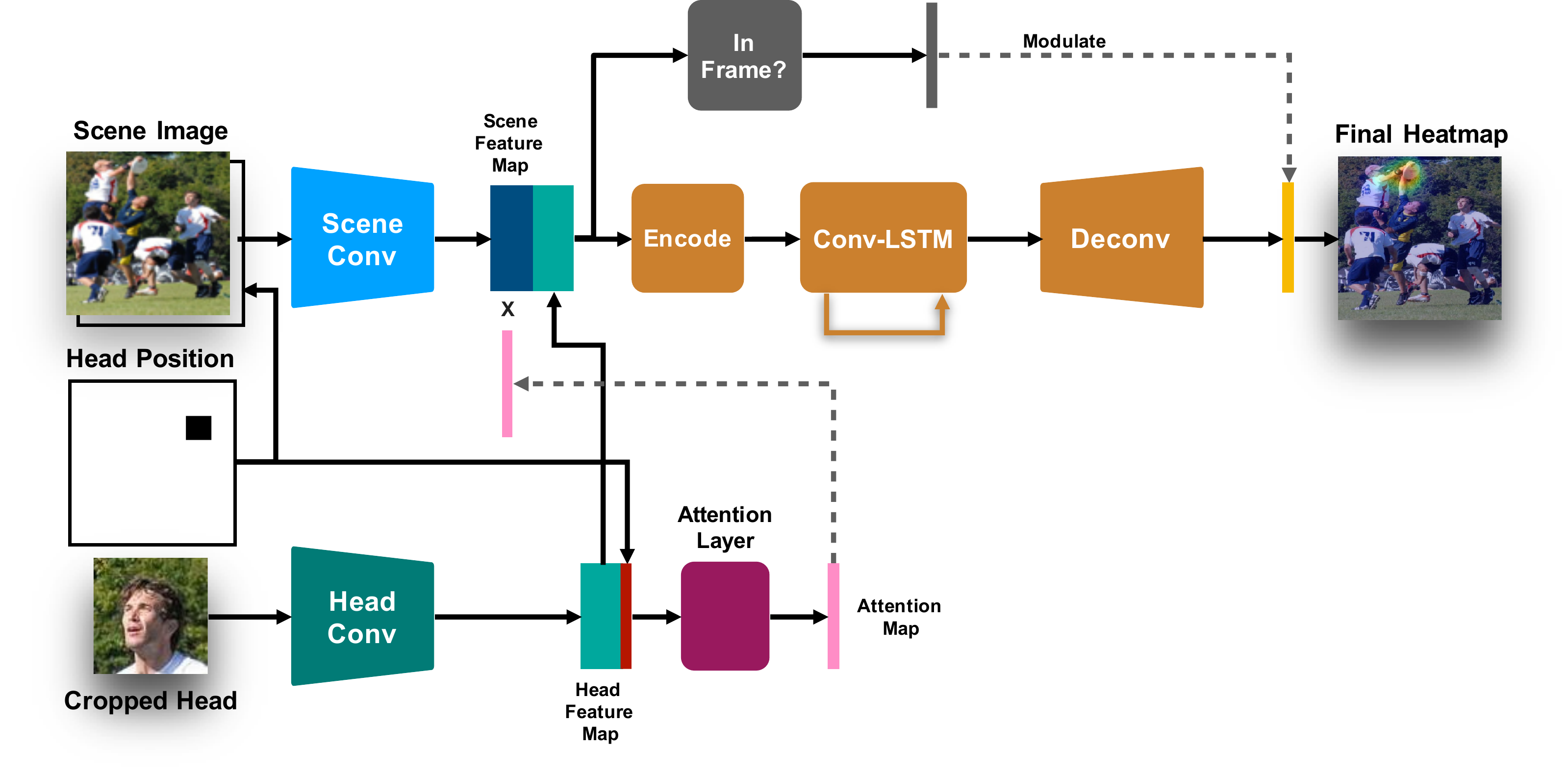}
    \caption{\textbf{Spatiotemporal architecture for gaze prediction.} It consists of a head conditioning branch which regulates the main scene branch using an attention mechanism. A recurrent module generates a heatmap that is modulated by a scalar, which quantifies whether the gaze target is in-frame. Displayed is an example of in-frame gaze from the GazeFollow dataset.}
    \label{fig:architecture}
    \vspace{-3mm}
\end{figure*}

\section{VideoAttentionTarget Dataset}
In this section we describe our novel \emph{VideoAttentionTarget} dataset that was created specifically for the task of video gaze target modeling. Some example frames, annotations and statistics of the dataset are shown in Fig~\ref{fig:dataset_showcase}. 

In order to ensure that our dataset reflects the natural diversity of gaze behavior, we gathered videos from various sources including live interviews, sitcoms, reality shows, and movie clips, all of which were available on YouTube. Videos from 50 different shows were selected. From each source video, we extracted short clips that contain dynamic gaze behavior without scene cuts, in which a person of interest can be continuously observed. The length of the clips varies between 1-80 seconds. 

For each clip, annotators first labeled tracks of head bounding boxes for each person. This resulted in 1,331 tracks comprising 164,541 frame-level bounding boxes. In the second pass, the annotators labeled the gaze target as a point in each frame for each annotated person. They also had the option to mark if the target was located outside the video frame (including the case where the subject was looking at the camera). This produced 109,574 in-frame gaze targets and 54,967 out-of-frame gaze annotations. All frames in all clips were annotated using custom software by a team of four annotators, with each frame annotated once. 


A testing set was constructed by holding out approximately 20\% of the annotations (10 shows, 298 tracks, 31,978 gaze annotations), ensuring no overlap of shows between the train and test splits. This allows us to measure generalization to new scenarios and individuals. Furthermore, in order to characterize the variability in human annotations of gaze targets, we had two other annotators (among the four who annotated the train split) who did not label that particular test samples additionally annotate them. We report this human inter-rater reliability which serves as the upper bound on the algorithm performance. 

\section{Spatiotemporal Gaze Architecture}
\label{lab:architecture}

Our architecture is composed of three mains parts. A \textbf{head conditioning branch}, a \textbf{main scene branch} and a \textbf{recurrent attention prediction module}. An illustration of the architecture is shown in Fig.~\ref{fig:architecture}.

\textbf{Head Conditioning Branch} The head conditioning branch computes a head feature map from the crop of the head of the person of interest in the image. The ``Head Conv'' part of the network is a ResNet-50~\cite{he2016deep} followed by an additional residual layer and an average pooling layer. A binary image of the head position, with black pixels designating the head bounding box and white pixels on the rest of the image, is reduced using three successive max pooling operations and flattened. We found that the binary image encoded the location and relative depth of the head in the scene more effectively than the position encoding used in previous works. The head feature map is concatenated with this head position feature. An attention map is then computed by passing these two concatenated features through a fully-connected layer which we call the ``Attention Layer''.

\textbf{Main Scene Branch} A scene feature map is computed using the ``Scene Conv'' part of the network, which is identical to the ``Head Conv'' module previously described. Input to the ``Scene Conv'' is a concatenation of scene image and head position image. We found that providing head position as a spatial reference along with the scene helped the model learn faster. This scene feature map is then multiplied by the attention map computed by the head conditioning branch. This enables the model to learn to pay more attention to the scene features that are more likely to be attended to, based on the properties of the head. In comparison to~\cite{chong2018connecting}, our approach results in earlier fusion of the scene and head information. The head feature map is additionally concatenated to the weighted scene feature map. Finally, the concatenated features are encoded using two convolutional layers in the ``Encode'' module.

\textbf{Recurrent Attention Prediction Module} After encoding, the model integrates temporal information from a sequence of frames using a convolutional Long Short-Term Memory network~\cite{xingjian2015convolutional}, designated as ``Conv-LSTM'' in Fig.~\ref{fig:architecture}. A deconvolutional network comprised of four deconvolution layers, designated as the ``Deconv'' module, upsamples the features computed by the convolutional LSTM into a full-sized feature map. We found that this approach yields finer details than the grid-based map used in~\cite{chong2018connecting}.

\textbf{Heatmap Modulation} The full-sized feature map is then modulated by a scalar $\alpha$ which quantifies whether the person's focus of attention is located inside or outside the frame, with higher values indicating in-frame attention. This $\alpha$ is learned by the ``In Frame?'' module in Fig.~\ref{fig:architecture}, which consists of two convolutional layers followed by a fully-connected layer. The modulation is performed by an element-wise subtraction of the $(1-\alpha)$ from the normalized full-sized feature map, followed by clipping of the heatmap such that its minimum values are $\geq 0$. This yields the final heatmap which quantifies the location and intensity of the predicted attention target in the frame. In Fig.~\ref{fig:architecture} we overlay the final heatmap on the input image for visualization.

\textbf{Implementation Details} We implemented our model in PyTorch. The input to the model is resized to 224$\times$224 and normalized. The Attention Layer outputs 7$\times$7 spatial soft-attention weights. The ConvLSTM module uses two ConvLSTM layers with kernels of size 3, whose output is up-sampled to a 64$\times$64-sized heatmap. Further model specifications can be found in our code.

For supervision, we place a Gaussian weight around the center of the target to create the ground truth heatmap. Heatmap loss $\mathcal{L}_{h}$ is computed using MSE loss when the target is in frame per ground truth. In-frame loss $\mathcal{L}_{f}$ is computed with binary cross entropy loss. Final loss $\mathcal{L}$ used for training is a weighted sum of these two: 
$\mathcal{L} = w_{h} \cdot \mathcal{L}_{h} + w_{f} \cdot \mathcal{L}_{f}$.

We initialize the Scene Conv with CNN for scene recognition~\cite{zhou2014learning} and the Head Conv with CNN for gaze estimation~\cite{funes2014eyediap}. Training is performed in a two-step process. First, the model is globally trained on the GazeFollow dataset until convergence. Second, it is subsequently trained on the VideoAttentionTarget dataset, while freezing the layers up to the Encode module to prevent overfitting. We used random flip, color jitter, and crop augmentations as described in~\cite{recasens2015they}. We also added noise to head position during training to minimize the impact of head localization errors.

\begin{table}[t]
    \begin{center}
        \begin{tabular}{lccc}
        \hline
        Method        & AUC $\uparrow$ & Average Dist. $\downarrow$ & Min Dist. $\downarrow$\\
        \hline
        \hline
        Random  & 0.504  & 0.484 & 0.391 \\ 
        Center  & 0.633  & 0.313 & 0.230 \\
        Judd~\cite{judd2009learning}  & 0.711  & 0.337 & 0.250 \\
        GazeFollow~\cite{recasens2015they}   & 0.878  & 0.190 & 0.113 \\
        Chong~\cite{chong2018connecting}  & 0.896  & 0.187 & 0.112 \\
        Zhao~\cite{zhao2019learning} & n/a  & 0.147 & 0.082 \\
        Lian~\cite{lian2018believe} & 0.906  & 0.145 & 0.081 \\
        Ours  & \textbf{0.921}  & \textbf{0.137} & \textbf{0.077} \\
        \hline
        \hline
        Human  & 0.924  & 0.096 & 0.040 \\
        \hline
        \end{tabular}
        \caption{\textbf{Spatial module evaluation} on the GazeFollow dataset for single image gaze target prediction.}\label{tab:eval_gazefollow}
        \vspace{-5mm}
    \end{center}
\end{table}

\begin{table}[t]
    \begin{center}
        \begin{tabular}{lcccc}
        \hline
        \multirow{2}{*}{Method} &  \multicolumn{2}{c}{\textit{spatial}} & \multicolumn{1}{c}{\textit{out of frame}}\\
          & AUC $\uparrow$ & $L^{2}$ Dist. $\downarrow$  & AP $\uparrow$\\
        \hline
        \hline
        Random  & 0.505 & 0.458 & 0.621\\
        Fixed bias & 0.728 & 0.326 & 0.624\\
        Chong~\cite{chong2018connecting}  & 0.830 & 0.193 & 0.705\\
        Chong~\cite{chong2018connecting}+LSTM & 0.833 & 0.171 & 0.712\\
        \hline
        No head position & 0.835 & 0.169 & 0.827\\
        No head features & 0.758 & 0.258 & 0.714\\
        No attention map & 0.717 & 0.226 & 0.774\\
        No fusion & 0.853 & 0.165 & 0.817\\
        No temporal & 0.854 & 0.147 & 0.848\\
        \hline
        Ours full & \textbf{0.860} & \textbf{0.134} & \textbf{0.853}\\
        \hline
        \hline
        Human & 0.921 & 0.051 & 0.925\\
        \hline
        \end{tabular}
        \caption{\textbf{Quantitative model evaluation} on our VideoAttentionTarget dataset.}\label{tab:main_quantitative}
        \vspace{-5mm}
    \end{center}
\end{table}

\section{Experiments}
\label{sec:exp}

We conducted four experiments to evaluate the performance of our method. Sec.~\ref{sec:spatial} uses just the spatial component of our model on the GazeFollow dataset. Sec.~\ref{sec:spatiotemp} uses the full spatiotemporal model on the VideoAttentionTarget dataset. Sec.~\ref{sec:app1} uses the model output to classify clinically-relevant social behaviors in a sample of toddlers. Sec.~\ref{sec:app2} uses the model to detect shared attention in the VideoCoAtt dataset. \emph{Our method produces state-of-the-art results on all datasets in all experiments.}

\begin{figure}[t]
    \centering
    \includegraphics[width=1\linewidth]{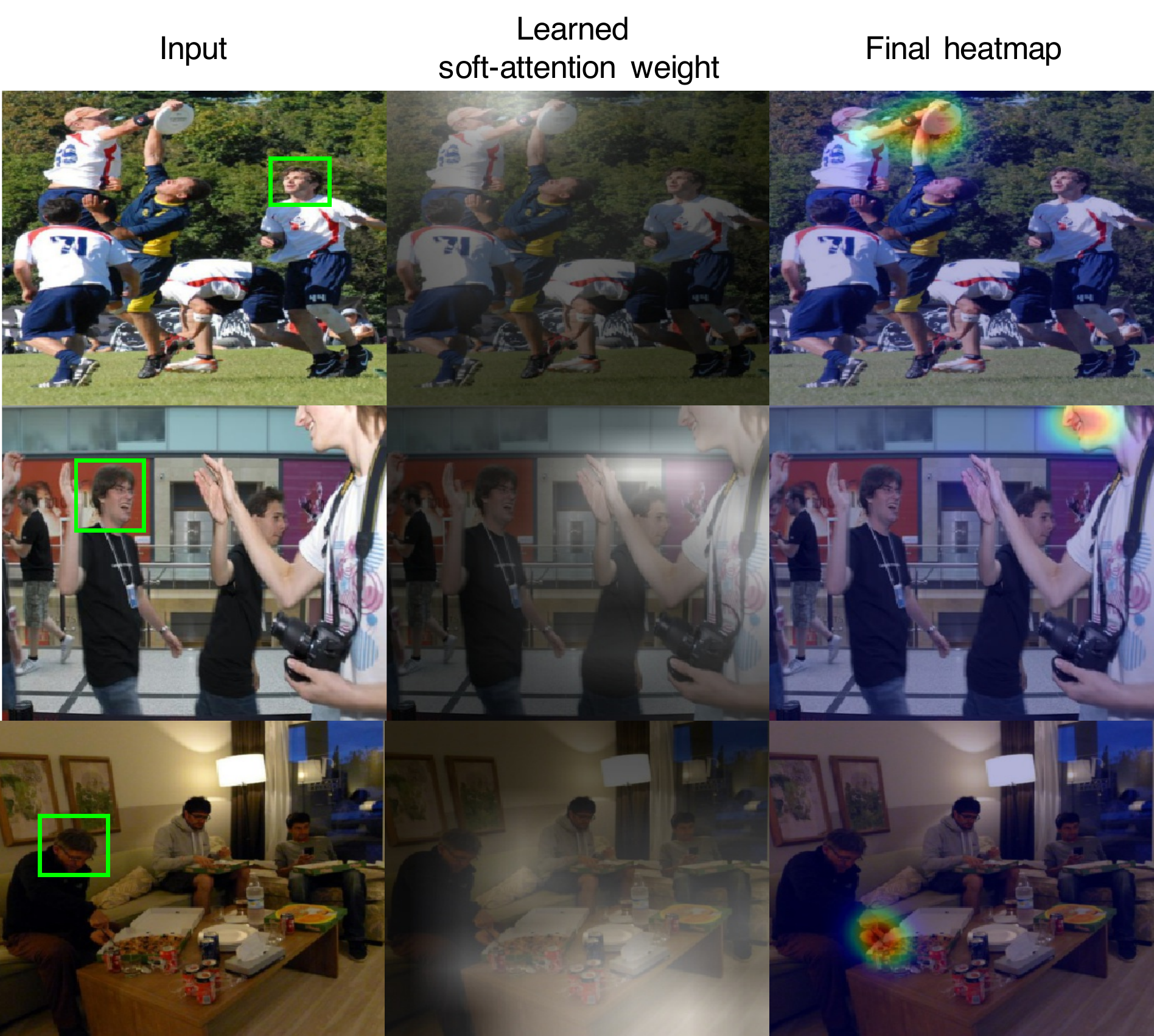}
    \caption{\textbf{Visualization of head-conditioned attention} with corresponding input and final output. The attention layer captures and leverages the head pose information to regulate the model's prediction.}
    \label{fig:example_attn_mech}
    \vspace{-3mm}
\end{figure}

\subsection{Spatial Module Evaluation}
\label{sec:spatial}

We evaluate the static part of our model on single image gaze target prediction using the GazeFollow dataset~\cite{recasens2015they}, and compare against prior methods. Evaluation follows the same protocol from~\cite{recasens2015they,chong2018connecting}. GazeFollow contains annotations of person heads and gaze locations in a diverse set of images. To train the model, we used the annotation labels from \cite{chong2018connecting} which were extended to additionally specify whether the annotated gaze target is out-of-frame. In order to make a fair comparison, we only use the GazeFollow dataset for training and do not include our new dataset in this experiment. 

The results in Table~\ref{tab:eval_gazefollow} demonstrate the value of our architectural choices in the spatial model component. We outperform previous methods by a significant margin. In fact, our AUC of 0.921 is quite close to the AUC of 0.924 obtained by human.
Qualitatively, visualization of the learned weights of the attention layer reveals that the model has learned to effectively make use of the facial orientation information for weighting scene features, as shown in Fig.~\ref{fig:example_attn_mech}.

\begin{figure}[ht]
    \centering
    \includegraphics[width=1\linewidth]{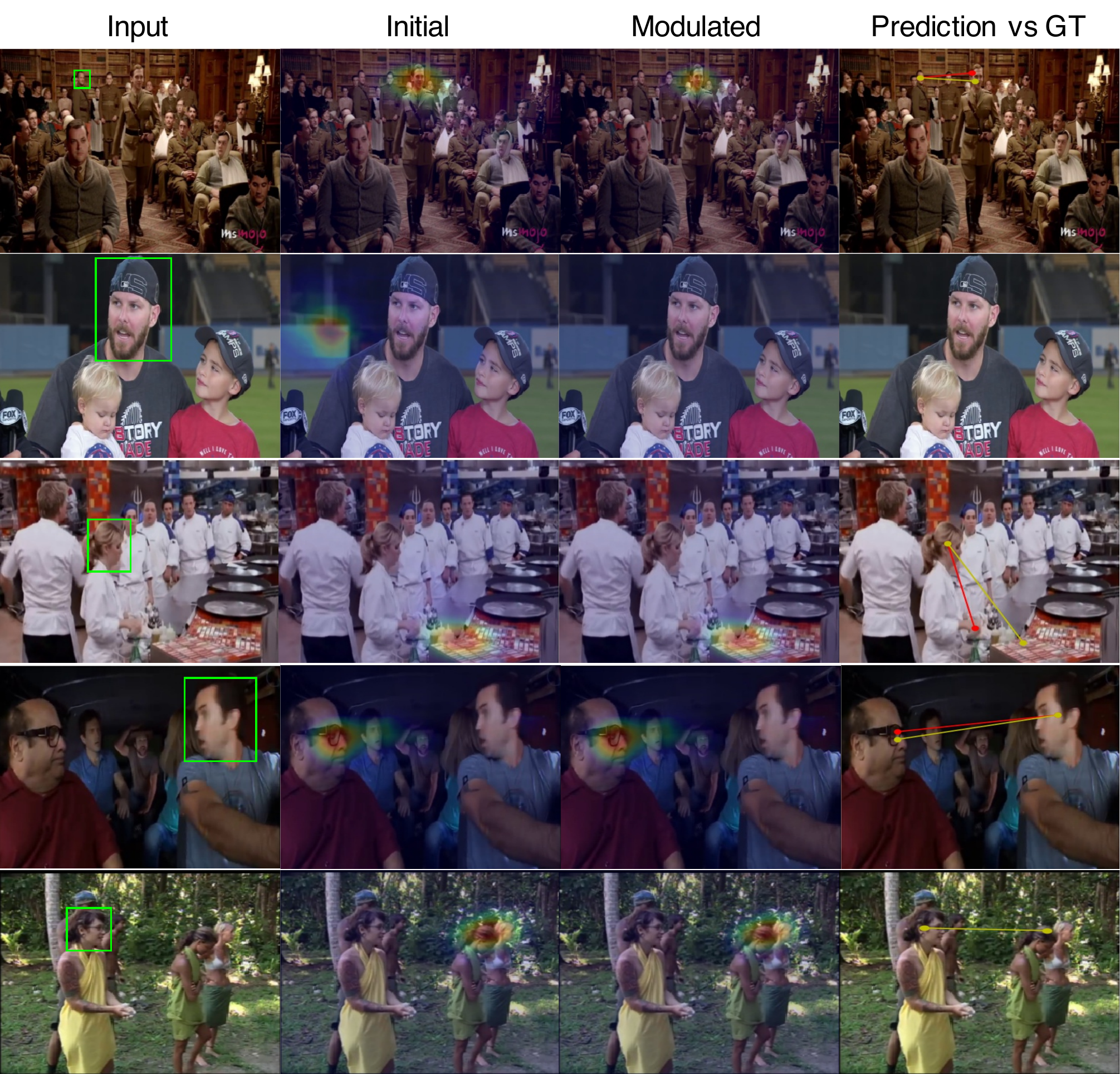}
    \caption{\textbf{Gaze target prediction results on example frames.} \textit{Initial} denotes the first output of the deconvolution, \textit{Modulated} shows the adjusted heatmap after modulation. Final prediction (yellow) and ground truth (red) are presented in the last column. Rows 1, 3, 4 depict properly predicted within-image gaze target, row 2 shows correctly identified nonexistent gaze target in frame, and the last row is an example of failure case where it predicts a fixated target behind the subject in the image due to the lack of sense of depth.}
    \label{fig:qualitative_ours}
    \vspace{-3mm}
\end{figure}

\subsection{Spatiotemporal Model Evaluation}
\label{sec:spatiotemp}
We evaluate our full model on the new \emph{VideoAttentionTarget} dataset. We use three performance measures: AUC, Distance, and Out-of-Frame AP. \textbf{AUC:} Each cell in the spatially-discretized image is classified as gaze target or not. The ground truth comes from thresholding a Gaussian confidence mask centered at the human annotator's target location. The final heatmap provides the prediction confidence score which is evaluated at different thresholds in the ROC curve. The area under curve (AUC) of this ROC curve is reported. \textbf{Distance:} $L^{2}$ distance between the annotated target location and the prediction given by the pixel of maximum value in the heatmap, with image width and height normalized to 1. AUC and Distance are computed whenever there is an in-frame ground truth gaze target (the heatmap always has a max). \textbf{Out-of-Frame AP:} The average precision (AP) is computed for the prediction score from the scalar $\alpha$ (described in Sec.~\ref{lab:architecture}) against the ground truth, computed in every frame. 
We also evaluate the performance of the annotators (\textbf{Human} performance) across all three measures. This is done by comparing annotator predictions in all pairs and averaging them. This is analogous to the kappa score used to measure inter-rater reliability, but specialized for our performance measures. 

Table~\ref{tab:main_quantitative} summarizes the experimental results. The first block of rows shows baseline tests and comparison with previous methods; \textit{Random} is when the prediction is made at 50\% chance, and \textit{Fixed bias} is when the bias present in the dataset (Fig~\ref{fig:dataset_stat}) is utilized. The method of~\cite{chong2018connecting}, which is the existing non-temporal gaze target estimator, is compared both as-is and using an additional LSTM layer on top. The second set of rows in the table shows ablation study results by disabling key components of our model one at a time; \textit{No head position} is when the head position image is not used. \textit{No head features} is when the head feature map from the Head Conv module is not provided. In this case, the attention map is made using only the head position. \textit{No attention map} is when attention map is not produced therefore the scene feature map is uniformly weighted. \textit{No fusion} is when the head feature map is only used to produce attention map and not concatenated with scene feature map for encoding. \textit{No temporal} is when ConvLSTM is not used. This quantitative analysis demonstrates that our proposed model strongly outperforms previous methods as well as the presented baselines. All components of the model are crucial to achieving the best performance, and the head convolutional pathway and the attention mechanism were found to have the biggest contribution. Qualitative results are presented in Fig.~\ref{fig:qualitative_ours}.

\subsection{Detecting the Social Bids of Toddlers}
\label{sec:app1}

\textbf{Motivation }
Eye contact and joint attention are among the earliest social skills to emerge in human development~\cite{carpenter1998social}, and are closely-related to language learning~\cite{hirotani2009joint} and socio-emotional development~\cite{macpherson2017attentional}. 
Children with autism exhibit difficulty in modulating gaze during social interactions~\cite{mundy1986defining, charman1997infants, senju2009atypical}, and social gaze is assessed as part of the diagnosis and treatment of autism. This is usually done qualitatively or through laborious manual methods. The ability to automatically quantify children's social gaze would offer significant benefits for clinicians and researchers.

\textbf{Toddler Dataset}
We sampled a dataset of 20 toddlers from~\cite{chong2017detecting} (10 with an autism diagnosis, 10 female, mean age 36.4 months) who were video-recorded during dyadic social interactions. In this dataset, each participant completed an assessment known as the ESCS~\cite{mundy2003early}, which was administered by trained examiners. The ESCS is a semi-structured play protocol designed to elicit nonverbal social communication behaviors.

Five expert raters annotated all of the child's gaze behavior consisting of looks to toys and looks to the examiner's face at the frame level. Based on this per-frame annotation, a toy-to-eyes gaze shift event is inferred if the gaze target changes from the toy to the examiner's face within 700 milliseconds. In total, the dataset contains 623 shift events during 221-minute-long recordings. Our task was to detect these toy-to-eyes gaze shifts and reject all other types of gaze shifts which the child made during the session. The toy-to-eyes shifts are relevant to child development because they can be further classified into different types of joint attention based on the context in which they are produced~\cite{mundy2003early}. Joint attention is a key construct for the development of social communication. Our experiment provides preliminary evidence for the feasibility of automatically identifying such gaze-based joint attention events from video.

\begin{figure}[t]
    \centering
    \includegraphics[width=1\linewidth]{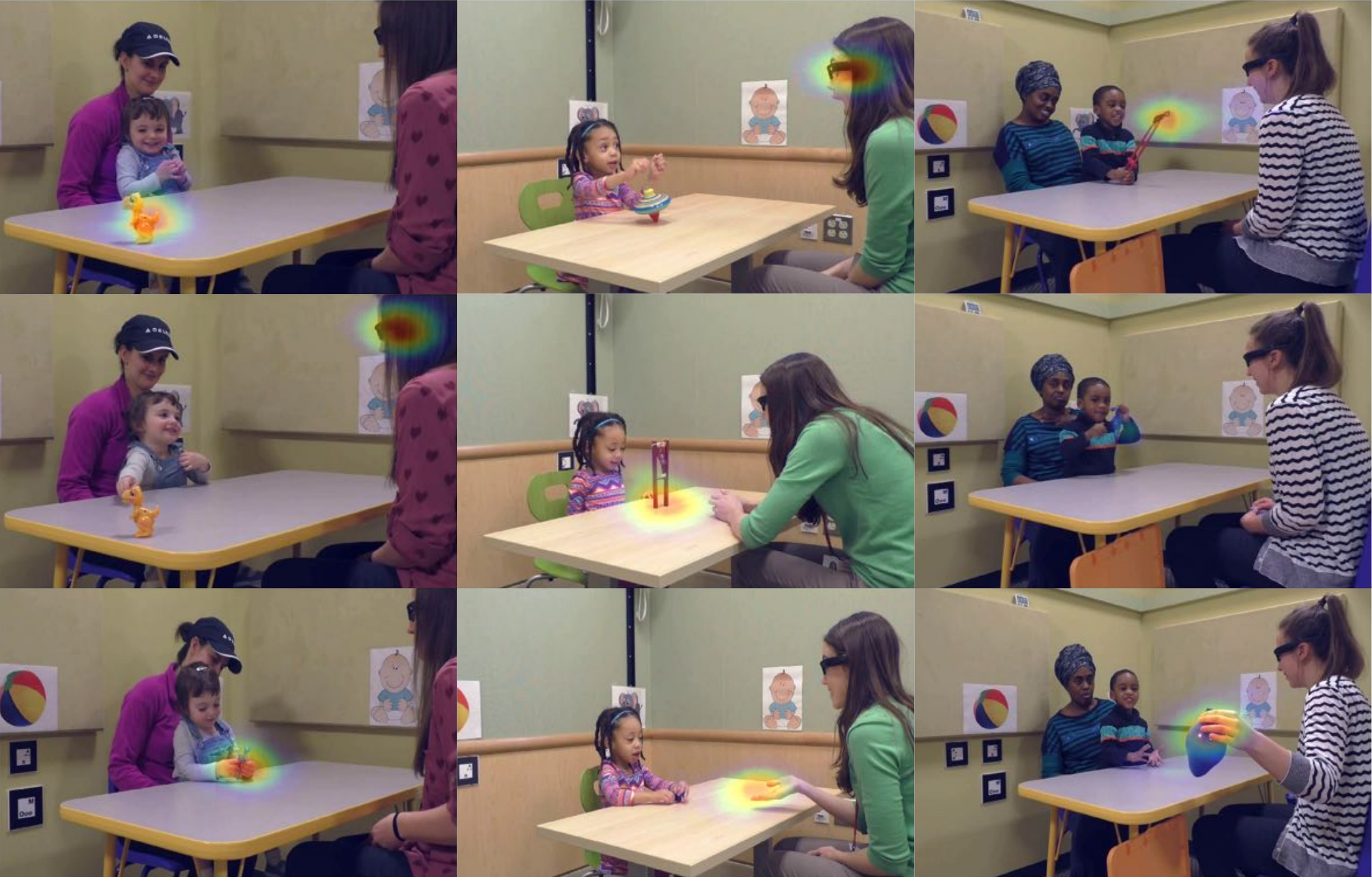}
    \caption{\textbf{Heatmap output} of our model on toddlers video.}
    \label{fig:shift_maps}
    \vspace{-3mm}
\end{figure}

\textbf{Experimental setup and results }
Given the toddlers dataset, we conducted experiments to see how an automated method can be used to retrieve gaze shift events. Two types of approaches for shift detection are explored. The first approach is to detect a shift in a two-step process where we initially classify the type of attended object - among toy, eyes, and elsewhere - in every frame with a ResNet-50 image classifier, and then apply an event classifier on top of it over a temporal window to conclusively find the gaze shift from toy to eyes. A random forest model is used for the event classifier. For the second approach, we try detecting a shift event in an end-to-end manner, using the I3D model~\cite{carreira2017quo} since gaze shift can be viewed as a special case of a human action. 

For both approaches, we compare shift detection performance when the inputs to the models are 1. the RGB image alone, 2. image and head position, and 3. image and heatmap produced by our attention network (Fig.~\ref{fig:shift_maps}). For 2 and 3 the head position or heatmap is concatenated depth-wise to the RGB image as a 4th channel in grayscale. CNN layers of ResNet were pretrained on ImageNet~\cite{deng2009imagenet} and those of I3D were pretrained on Kinetics~\cite{kay2017kinetics}. The child's head was detected and recognized using~\cite{dlib09}. A sliding window size of 64 frames was used during training. For validation, we adopted 5-fold subject-wise cross validation in which 4 subjects were held out in each validation set.

Table~\ref{tab:simons_table} summarizes the results of our experiment with the precision and recall of gaze shift detection. Interestingly, the 2D-CNN-based approach generally outperformed the 3D-CNN method, which is presumably due to the complexity of I3D and relatively less training data. 
Nevertheless, there still exists a noticeable gap relative to human performance, implying the need for further research on this problem.

\begin{table}[ht]
    \begin{center}
        \begin{tabular}{l|c|cc}
        \hline
        \multirow{2}{*}{Method} & Detection & \multirow{2}{*}{Prec. $\uparrow$} & \multirow{2}{*}{Rec. $\uparrow$}\\
          &  Approach &  &  \\
        \hline
        \hline
        Random  &  & 0.034 & 0.503 \\
        \hline
        ResNet on RGB & \multirow{2}{*}{random} & 0.541 & 0.567\\
        ResNet on RGB+head & \multirow{2}{*}{forest} & 0.598 & 0.575\\
        ResNet on RGB\textbf{+hm} &  & \textbf{0.708} & \textbf{0.759}\\
        \hline
        I3D on RGB & \multirow{3}{*}{end-to-end}  & 0.433 & 0.506\\
        I3D on RGB+head &   & 0.475 & 0.500\\
        I3D on RGB\textbf{+hm} & & 0.559 & 0.710\\
        \hline
        \hline
        Human (clinical experts) &  & {0.903} & {0.922} \\
        \hline
        \end{tabular}
        \caption{\textbf{Gaze coding detection results} on the toddlers dataset. As shown, our heatmap feature (denoted as \textbf{hm}) indeed improves shift detection when used along with image in a standard classification paradigm.}\label{tab:simons_table}
        \vspace{-3mm}
    \end{center}
\end{table}

\subsection{Detecting Shared Attention in Social Scenes}
\label{sec:app2}

As an additional application of our system on real-world problems, we apply our model to infer shared attention in social scenes. We use the VideoCoAtt dataset~\cite{fan2018inferring} to benchmark our performance on this task. This dataset has 113,810 test frames that are annotated with the target location when it is simultaneously attended by two or more people. 

Given that our model does not have a head detection module as in~\cite{fan2018inferring}, we trained a SSD-based~\cite{liu2016ssd} head detector in the same manner as~\cite{marin2019laeo} to automatically generate the input head positions. We fine-tuned this head detector with the head annotations in VideoCoAtt. However, we chose not to fine-tune our model for gaze target detection with VideoCoAtt, since their annotations do not naturally translate to the dense single-subject-target annotations that our model requires for training.

\begin{figure}[t]
    \centering
    \includegraphics[width=1\linewidth]{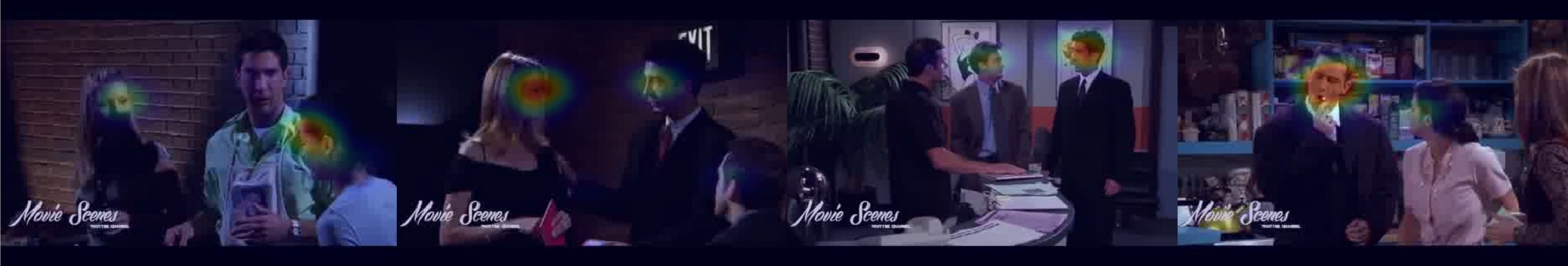}
    \caption{\textbf{Constructed shared attention map} obtained by adding up individual heatmaps of all people in the image. Samples are from the VideoCoAtt dataset.}
    \label{fig:coatt_map}
\end{figure}

Our method is evaluated on the following two tasks: 1. location prediction (spatial) and 2. interval detection (temporal) of shared attention. For the localization task, we first add up the individual heatmaps of all people in the frame and aggregate them into a single shared attention confidence map (examples in Fig.~\ref{fig:coatt_map}). Then, the $L^{2}$ distance is computed between the pixel location of the maximum confidence and the center of the ground truth. 
For the interval detection task, we regard the aggregated confidence map as representing a shared attention case if its maximum score is above certain threshold. A single heatmap from our model can have a maximum value of 1 at the fixated location and when another heatmap is added to the same location its value becomes 2. We chose a threshold value of 1.8 instead of 2 in this experiment to make a room for slight misalignments between multiple fixations.

As a result, our method achieves state-of-the-art results on both tasks, as shown in Table~\ref{tab:eval_coatt}. This outcome is surprising since the models of~\cite{fan2018inferring,Sumer_2020_WACV} were formulated specifically to detect shared attention, whereas ours was not. However, it must also be noted that there exist differences in the experimental setup, such as the head detector and the training data, thus there are some caveats associated with our experimental finding. Here, we intend to demonstrate the potential value of our model for recognizing higher-level social gaze, and it is encouraging that we can achieve good performance without tweaking the model for this specific problem.

\begin{table}[t]
    \begin{center}
        \begin{tabular}{lcc}
        \hline
        Method        & Accuarcy $\uparrow$ & $L^{2}$ Dist. $\downarrow$ \\
        \hline
        \hline
        Random  & 50.8 & 286 \\ 
        Fixed bias  & 52.4  & 122 \\
        GazeFollow~\cite{recasens2015they} & 58.7 & 102 \\
        Gaze+Saliency~\cite{pan2016shallow} & 59.4 & 83 \\
        Gaze+Saliency~\cite{pan2016shallow}+LSTM & 66.2 & 71 \\
        Fan~\cite{fan2018inferring} & 71.4 & 62 \\
        Sumer~\cite{Sumer_2020_WACV} & 78.1 & 63 \\
        Ours & \textbf{83.3} & \textbf{57} \\
        \hline
        \end{tabular}
        \caption{\textbf{Shared attention detection results} on the VideoCoAtt dataset. The interval detection task is evaluated with prediction accuracy and the localization task is measured with $L^{2}$.}\label{tab:eval_coatt}
        \vspace{-6mm}
    \end{center}
\end{table}

\section{Conclusion}
We have presented a new deep architecture and a novel \emph{VideoAttentionTarget} dataset for the task of detecting the time-varying attention targets for each person in a video. Our model is designed to allow the face to direct the learning of gaze-relevant scene regions, and our new dataset makes it possible to learn the temporal evolution of these features. The strong performance of our method on multiple benchmark datasets and a novel social gaze recognition task validates its potential as a useful tool for understanding gaze behavior in naturalistic human interactions.

\section{Acknowledgement}
We thank Caroline Dalluge and Pooja Parikh for the gaze target annotations in the VideoAttentionTarget dataset, and Stephan Lee for building the annotation tool and performing annotations. The toddler dataset used in Sec.~\ref{sec:app1} was collected and annotated under the direction of Agata Rozga, Rebecca Jones, Audrey Southerland, and Elysha Clark-Whitney. This study was funded in part by the Simons Foundation under grant 383667 and NIH R01 MH114999.

{\small
\bibliographystyle{ieee_fullname}
\bibliography{bib}
}

\end{document}